# Autonomous Learning of Features for Control: Experiments with Embodied and Situated Agents

**Nicola Milano and Stefano Nolfi**
Institute of Cognitive Science and Technologies
National Research Council (CNR-ISTC)
Roma, Italy
stefano.nolfi@, nicola.milano@istc.cnr.it

**Abstract**
As discussed in previous studies, the efficacy of evolutionary or reinforcement learning algorithms for continuous control optimization can be enhanced by including a neural module dedicated to feature extraction trained through self-supervised methods. In this paper we report additional experiments supporting this hypothesis and we demonstrate how the advantage provided by feature extraction is not limited to problems that benefit from dimensionality reduction or that involve agents operating on the basis of allocentric perception. We introduce a method that permits to continue the training of the feature-extraction module during the training of the policy network and that increases the efficacy of feature extraction. Finally, we compare alternative feature-extracting methods and we show that sequence-to-sequence learning yields better results than the methods considered in previous studies.

**1. Introduction**

The discovery of expressive features constitutes a critical pre-requisite for the development of effective control policies. In the case of robots, feature learning can be especially challenging as it involves the generation of features describing the state of the agent and of the environment from raw and noisy sensor measurements that provide only part of the required information.

Recent works in evolutionary (Salimans et al., 2017; Pagliuca et al. 2020) and reinforcement learning methods (Duan et al., 2016; Andrychowicz et al., 2019) demonstrated how adaptive robots can successfully learn effective policies directly from observation on the basis of reward signals that only rate the performance of the agent. In other words, these methods can discover the features required to realize effective control policies automatically without the need of additional dedicated mechanisms or processes. These approaches are usually referred with the term end-to-end learning. On the other hand, extended methods incorporating mechanisms or processes dedicated to the extraction of useful features can potentially speed-up learning and scale-up better to complex problems.

Feature learning is a general domain which aims to extract features that can be used to characterized data. Recently, this area achieved remarkably results in the context of neural network learning for classification and regression problems. For example, feature learning with deep convolutional neural networks constitutes an excellent solution for image classification problems. Autonomous learning of features for control is a special area of feature learning in which the input vectors (*observations*) are influenced by output vectors (*actions*) and in which observations and actions vary during the course of the learning process (Bohmer et al., 2015; Lesort et al., 2018). The term autonomous refers to the fact that the features are extracted through a self-supervised learning process, i.e. a learning process in which the supervision is provided directly by the data available as input.

The methods that can be used to learn features include auto-encoders, forward models, and cognitive biases or priors. Auto-encoders are trained to encode observations in more compact intermediate vectors that are then decoded to regenerate the original observation vectors. Forward models are trained to predict the next observation on the basis of the current observation and of the action that the agent is going to perform. The features extracted by auto-encoders or forward models can be provided as input to the policy network, trained through evolutionary or reinforcement learning algorithms, to maximize a task-dependent reward function. Cognitive biases or priors are constraints that promote the development of features characterized by certain desired properties in normal end-to-end learning, for example features fluctuating slowly over time (Lake et al., 2017; Bengio, Courville & Vincent, 2012; Lesort et al. 2017). Constraint of these types can be introduced by using cost functions that encourage the generation of features with the desired qualities.

A first objective of feature learning consists in generating compact representation of high dimensional sensory data which, in the case of robotic problems, might include 50 or more joint angles and velocities, hundreds of sensors encoding tactile information, and thousands or millions of pixels forming camera images. A second objective consists in generating useful features, i.e. features that are more informative than the observation states from which they can be extracted.

In this paper we focus on the second objective. More specifically, we analyze whether the features extracted through a self-supervised learning methods facilitate the development of effective solutions in embodied and situated robots trained through an evolutionary algorithm. Unlike previous related works, we consider experimental scenarios in which the robots have access to egocentric perceptual information.

Our results demonstrate how the usage of features extracted through auto-encoders, forward models, and sequence to sequence models (Srivastava et al. 2015, Sutskever et al. 2014, Cho et al. 2014) permits to achieve better solutions with respect to the end-to-end control condition. This providing that the self-supervised learning process, that determines the features extracted, is continued during the course of the evolutionary process. The comparison of the alternative methods that can be used to extract features indicates that the best results are obtained through the usage of sequence-to-sequence learning.

**2. Novelty and relation with related works**

Previous related works considered scenarios in which the agent can observe itself and the environment from an external perspective. In particular, Riedmiller and VoigtHinder (2012), considered the case of a racing slot car that can observe itself and the race track through a camera located above. The car is trained through reinforcement learning to move as fast as possible without crashing by receiving as input a low-dimensional representation of the scene extracted from a deep autoencoder neural network. This is realized in three phases. The first phase is used to generate the training set for the auto-encoder network that is created by collecting images from the camera at a constant frame rate while the car moves at a low constant speed on the track. The second phase is used to train the autoencoder network to learn an abstract compressed representation of the images (i.e. to compress the observations in a vector $z$ that is then used to regenerate the original observations). Finally, the third phase is used to train the policy network to drive the car. The network receives as input the vector $z$, extracted from the observations by the auto-associative network, and produce as output the desired speed of the car's wheels. A similar approach has been used to train an agent to swing a pendulum on the basis of visual information collected from a side view (Mattner, Lange & Riedmiller, 2012). The results reported by the authors demonstrated that the usage of the auto-encoder network permits to considerably speed-up learning with respect to an end-to-end approach.

More recently, Ha and Schmidhuber (2018) proposed a method that combines an auto-encoder network, a forward-model network, and a policy network. The model has been applied to the CarRacing-v0 problem (Klimov, 2016) that consists in driving a car on race tracks in simulation on the basis of images collected by a camera located above the track. Also in this case, the first phase is dedicated to the generation of the training set that is formed by the observations *o* experienced by the car and the actions *a* performed by the car during 10,000 episodes in which the car moves by selecting random actions. The second phase is used to train the auto-encoder network. The training enables the auto-encoded to extract an abstract compressed representation *z* of observations. During the third phase, the forward-model network is trained to predict $z_{t+1}$ on the basis $z_t$ and $a_t$. This enable the forward model to compress the observations of the agent over time in a vector *h*. Both the auto-encoder and the forward-model networks are trained on the basis of the training set collected in the first phase. Finally, in the fourth phase, the policy network is trained to drive the car so to maximize the cumulative rewards by receiving as input the *z* and *h* vectors extracted by the auto-associative and forward models. The policy network is trained by using the Covariance-Matrix Adaptation Evolution Strategy (CMA-ES) (Hansen & Ostermeier, 2001). The agents that receive as input the features extracted by the auto-associative and forward model achieve better performance than the agents that receive as input the observations.

In this work we investigate whether autonomous feature learning can be beneficial also in problems in which the agents receive egocentric observations, i.e. in problems in which the sensors of the agents are located on the agents' body. The availability of allocentric observations, collected from a camera located outside the agent, implies the necessity to transform the information contained in the observations to the agent's perspective, i.e. to the perspective that is relevant to decide the actions that the agent should perform. The need to perform this transformation might explain the advantage of feature extraction reported in the experiments reviewed above. By using egocentric observations, we can rule this possibility out. In other words, we can verify whether features extraction is advantageous in general, and not only in problems requiring a perspective transformation. For this reason, we selected classic benchmark problems that operate on the basis of egocentric perceptual information such as the BulletWalker2D, BulletHalfcheetah (Coumans and Bai, 2016), the BipedalWalkerHardcore (Brockman et al. 2016), and the MIT Racecar problem (https://racecar.mit.edu/, Coumans and Bai, 2016).

The size of the observation vector is relatively compact in these problems. This permits to verify whether feature extraction is advantageous in general and not only in problems that benefit from dimensionality reduction.

Another novelty with respect to previous study consists in: (i) the implementation and analysis of a method that permits to continue the training of the feature extracting network/s during the training of the policy network, and (ii) the utilization of sequence-to-sequence learning for feature extraction. As we will see, these two variations permit to further enhance the advantage provided by feature extraction.

## 3. Experimental setting

The Walker2DBullet (Figure 1, top-left) and HalfcheetahBullet (Figure 1, top-right) problems involve simulated robots with different morphologies rewarded for the ability to walk on a flat terrain (Coumans and Bai, 2016). The bullet version of this problems constitutes a free and more realistic implementation of the original MuJoCo problems (Todorov, Erez and Tassa, 2012). The robots have 6 actuated joints, respectively. The observation includes the position and orientation of the of the robot, the angular position and velocity of the actuated joints, and the contact sensors located on the feet. The initial posture of the robot varies randomly within limits. The evaluation episodes are terminated after 1000 steps or, prematurely, when the position of the torso of the robots falls below a threshold. The robots are rewarded on the basis of their velocity toward the target destination. In the case of the HalfcheetahBullet, the reward function includes also an additional component that punishes the agent with -0.1 for each joint currently located on one of the two limits. These rewards functions, optimized for evolutionary strategies, include fewer components than those optimized for reinforcement learning (Pagliuca, Milano and Nolfi, 2020).

The BipedalWalkerHardcore (Figure 1, bottom-left) is another walker problem that involves a two-legged 2D robot with 6 actuated joints situated in an environment including obstacles, e.g. ladders, stumps, and pitfalls. The robot is rewarded for moving forward and is penalized proportionally to the torque used. The robot's observations include the horizontal and vertical speed of the hull, the angular speed and velocity of the hull, the position and speed of the joints, the state of the contact sensors placed on the legs, and 10 proximity measures collected by a lidar rangefinder.

The MIT racecar problem involves a simulated version of the MIT racecar located on a race track (Figure 1, bottom-right). The car is provided with 2 actuators that control acceleration and steering. The observation includes 30 lidar proximity measures distributed over a range of 270 degrees on the frontal side of the car. The racetrack is divided in 110 consecutive virtual sections and the car is rewarded with 1 point every time it moves in a subsequent sector. The evaluation episodes are terminated after 10,000 steps or, prematurely, if the car remains stuck for 100 consecutive steps.

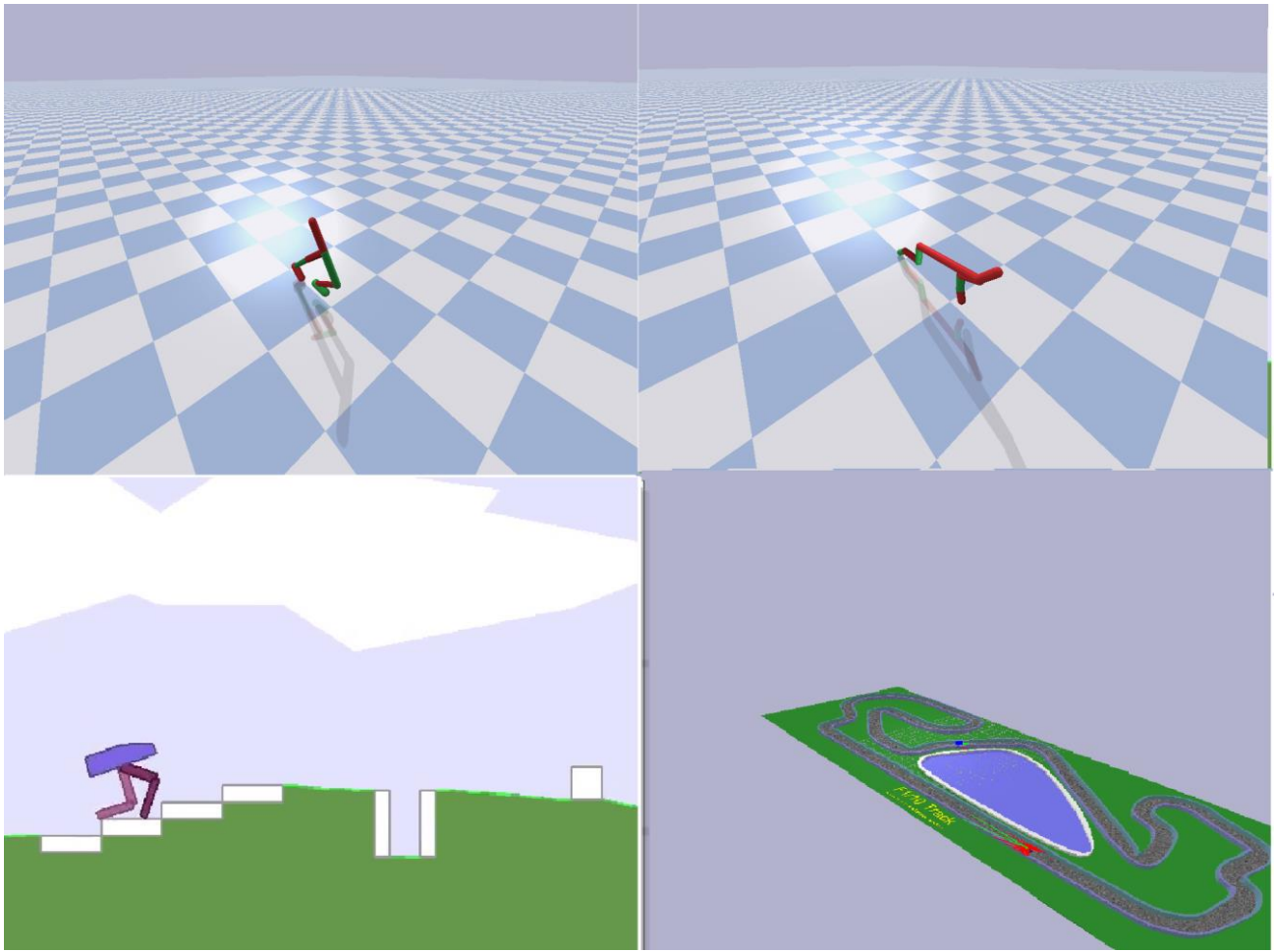

Figure 1. Illustration of the Walker2DBullet (top-left), HalfcheetahBullet (top-right), BipedalWalker (bottom-left), and MIT racecar (bottom-right) problems.

The robots are provided with a feed-forward neural network policies with an internal layer formed by 64 neurons. The number of input and output neurons depends on the size of the observation and action states and consists of [22, 26, 22, 30] inputs and [6, 6, 6, 2] outputs in the case of the Walker2dBullet, HafcheetahBullet, BipedalWalkerHardcore, and MIT racecar problems, respectively. The sensory neurons are updated on the basis of the state of the corresponding sensors. The internal and output neurons are updated with hyperbolic tangent and linear activation functions, respectively. The connection weights and the biases of the policy network are trained through the OpenAI evolutionary strategy (Salimans et al. 2017), i.e. one of the most effective evolutionary method for continuous problem optimization (Pagliuca, Milano and Nolfi, 2020). Episodes last 1,000 steps in the case of the Walker2DBullet, HalfCheetahBullet, and BipedalWalkerHardcore problems, and 10,000 steps in the case of the MIT racecar. Agents are evaluated for 1 episode in the case of the Walker2DBullet, HalfCheetahBullet, BipedalWalkerHardcore, and MIT racecar and for 5 episodes in the case of the BipedalWalkerHardcore. The usage of a greater number of episodes in the last case is due to the fact that the variability of the environmental conditions is greater. The training of the policy is continued for $20 * 10^7$ steps in the case of the BipedalWalkerHardcore problems and for $5 * 10^7$ steps in the case of the other problems.

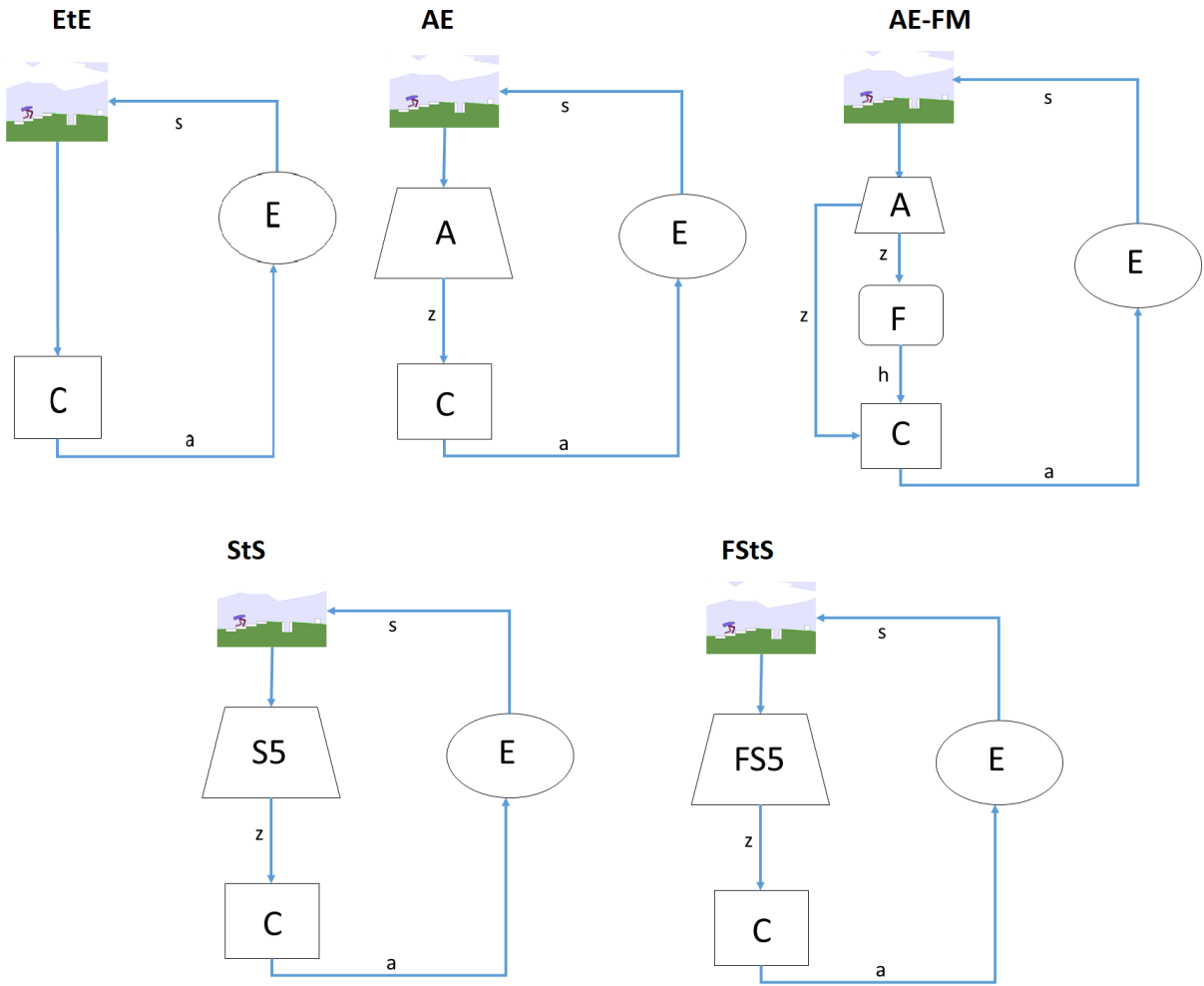

Figure 2. Schematization of the 5 experimental conditions (EtE, AE, AE-FM, StS, and FStS). The picture at the top of each image indicate the observation (e.g. the angle and velocity of the agent's joint, the proximity measures ext.). The action vectors are indicated with the letter *a*. The vector of features extracted from the auto-associative and forward model networks are indicated with the letter *z* and *h*, respectively. A indicates an auto-associative network that receives as input the observation at time *t* and produce as output the same observation at time *t*. F indicates a forward model network that receives as input the vector *z*, extracted from the auto-associative network at time *t*, and the action vector *a* at time *t* and produces as output the vector *z* at time $t_{+1}$. S5 indicates a sequence-to-sequence associative network that receives as input the observations at time $[t_{-4}, t]$ and produces as output the observation at time $[t_{-4}, t]$. FS5 indicates a sequence-to-sequence network that receives as input the observation at time $[t_{-4}, t]$ and produces as output the observations at time $[t_{-3}, t_{+1}]$. C indicates the control network (i.e. the policy network) that receives as input the observation (in the case of the EtE condition), or the features vector *z* (in the case of the AE, StS, and FStS conditions), or the features vectors *z* and *h* (in the case of the AE-FM condition). C produce as output the action vector that determines the movement of the robot's joints or wheels and consequently the state of the environment (E) at time $t_{+1}$. The new state of the environment determines the observation at time $t_{+1}$.

The different experimental conditions are illustrated in Figure 2.

In the end-to-end (**EtE**) condition the policy network receives as input directly the observation.

In the autoencoder (**AE**) condition the policy network receives as input the feature extracted from the observation by an autoencoder feed-forward neural network with 1 internal layer made of 50 neurons.

In the autoencoder forward-model condition (**AE-FM**) the policy network receives as input the features extracted by an auto-encoder network and by the forward model network. The former is constituted by a feed-forward network with 1 internal layer formed by 50 neurons. The latter is constituted by a LSTM network (Hochreiter and Schmidhuber, 1997) with 1 internal layer formed by 50 units. As in the Ha and Schmidhuber (2018) the forward-model network is trained to predict the internal state of the autoencoder $z$ at time $t_{+1}$ on the basis of the internal state of the autoencoder at time $t$ and of the action that the agent is going to perform.

In the sequence-to-sequence condition (**StS**) the policy network receives as input the features extracted by a sequence-to-sequence network that is trained to compress the last 5 observations into an internal vector $h$ that is used to regenerate the same observations.

Finally, in the forward sequence-to-sequence condition (**FStS**) the policy network receives as input the feature extracted from a sequence to sequence network that is trained to compress a set of 5 observations into an internal vector $h$ that is used to regenerate the last 4 observation and to predict the following future observation.

The training set used to train the auto-encoder, the forward model, and the sequence to sequence models is generated by collecting observation states obtained during 1000 episodes during which the agent performs random actions. The networks used to extract the feature are trained for 500 epochs. In the continuous training experiments, indicated with the symbol *, the training set is updated every generation by replacing the oldest 1% of the data with observations collected by agents acting on the basis of their current policy. The updated trained set is used to continue the training of the feature-extracting networks for 10 epochs every generation.

The experiments performed can be replicated by using the software available from https://github.com/milnico/features-extraction

## 4. Results

We first evaluated the five alternative models described above on the Walker2DBullet problem (Figure 2).

The top figure shows the results obtained in the experiments in which the training set used to train the feature extracting network/s is collected by agents performing random actions and in which the training is completed before the training of the policy network. The bottom figure, instead, shows the results in the experiments in which the training of the feature-extracting network/s is continued during the training of the policy network and in which the training set is updated with observations collected by agents acting on the basis of their current policy.

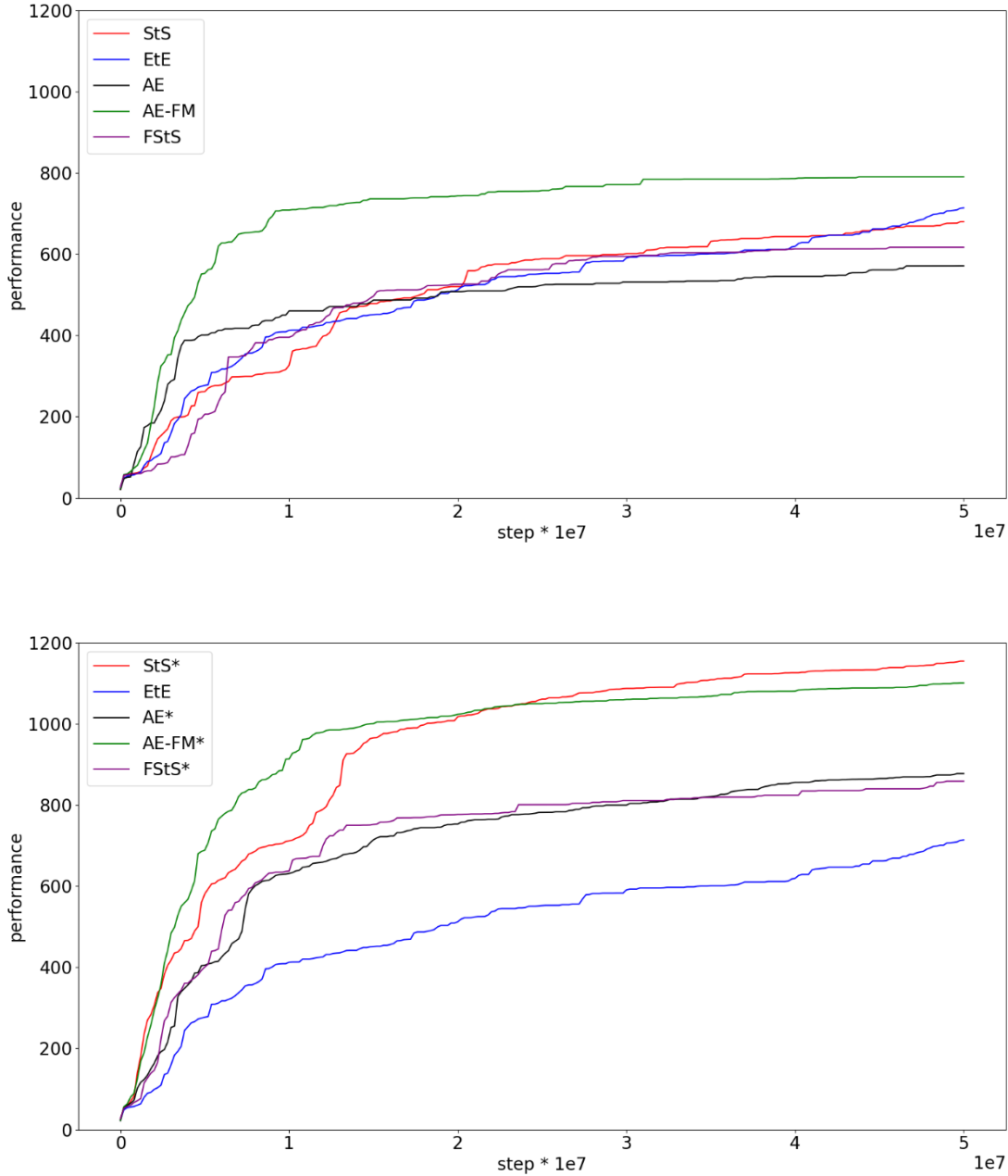

Figure 2. Performance during the training process in the case of the Walker2DBullet problem. Performance refer to the average fitness of the best agent achieved so far. Data computed by post-evaluating the best individual of each generation for 3 episodes. Each curve shows the average result of 10 replications. The top and the bottom figures show the results obtained in the experiments in which the training of the feature-extracting network/s is performed before the training of the policy network or is continued during the training of the policy network, respectively.

The experiments in which the training of the feature extracting network/s is continued during the training of the policy network outperform the experiments in which the training of the feature extracting network/s terminates before the training of the policy network in all cases (Figure 2 top and bottom, Mann-Whitney U test p-value < 0.01 in all conditions). This can be explained by considering that the observations experienced by performing random actions differ significantly from the observations experienced by agents trained to maximize the cumulative reward. In other words, the feature extracted from observation experienced after the execution of random actions do not necessarily capture the regularities that characterized the observations experienced by trained agents. This hypothesis is also supported by the fact the offset between the actual and desired outputs produced by the autoencoder, forward-model, sequence to sequence, and forward sequence to sequence networks increase considerably during the training of the policy network (in the experimental condition in which the feature-extracting network are trained before the policy network). On the contrary, when the training of the feature-extracting network is continued, the offset remains low (Figure 3, base versus * conditions).

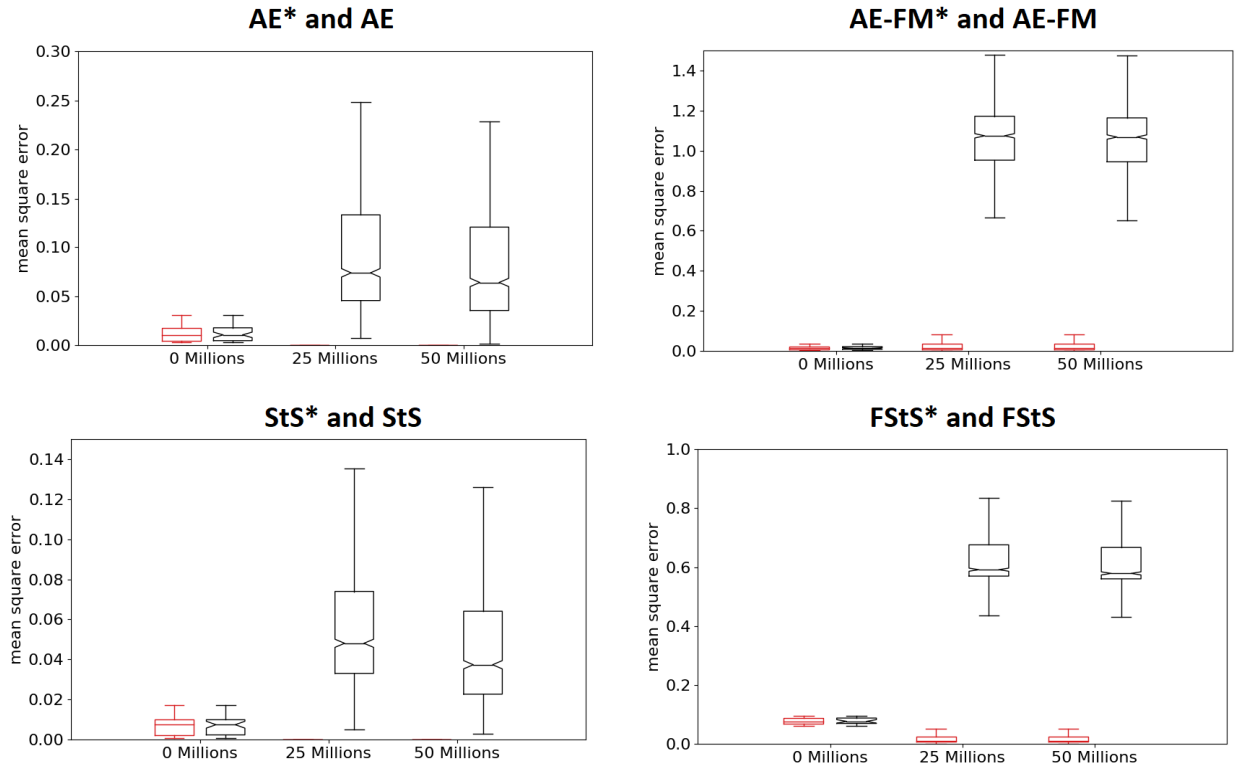

Figure 3. Mean-squared error produced by the features-extracting networks before the training of the policy network (0 steps) and during the training of the policy network (after 25 and 50 * $10^6$ steps). Results of the experiments carried out with the AE and AE*, AE-FM and AE-FM*, StS and StS*, and FStS and FStS* models. The boxplots shown in red and black shows the results obtained in the experiments in which the training of the feature extracting network is continued or not continued during the training of the policy network, respectively. In the case of AE* and StS* conditions, some of the red boxplots are not visible since the distribution of the offset is close to 0.0.

The analysis of the performance obtained by using different methods to extract features indicated that all methods achieve significantly better results than the end-to-end method (Mann-Whitney U test p-value < 0.01 in all conditions) in the experiments in which the training of the feature-extracting network is continued during the training of the policy. The comparison of the different methods indicate that the best results are obtained in the case of the StS* method that produces significantly better results than all alternative methods (Mann-Whitney U test p-value < 0.01).

We then compared the performance of the StS*and the EtE method on the other three problems considered (Figure 4). We focused on the StS* method since it achieved the best performance in the case of Walker2DBulletProblem. The StS* method outperforms EtE method in all case (Mann-Whitney U test p-value < 0.01) with the exception of the HalfcheetahBullet in which the performance in the two conditions do not differ statistically (Mann-Whitney U test p-value >0.01).

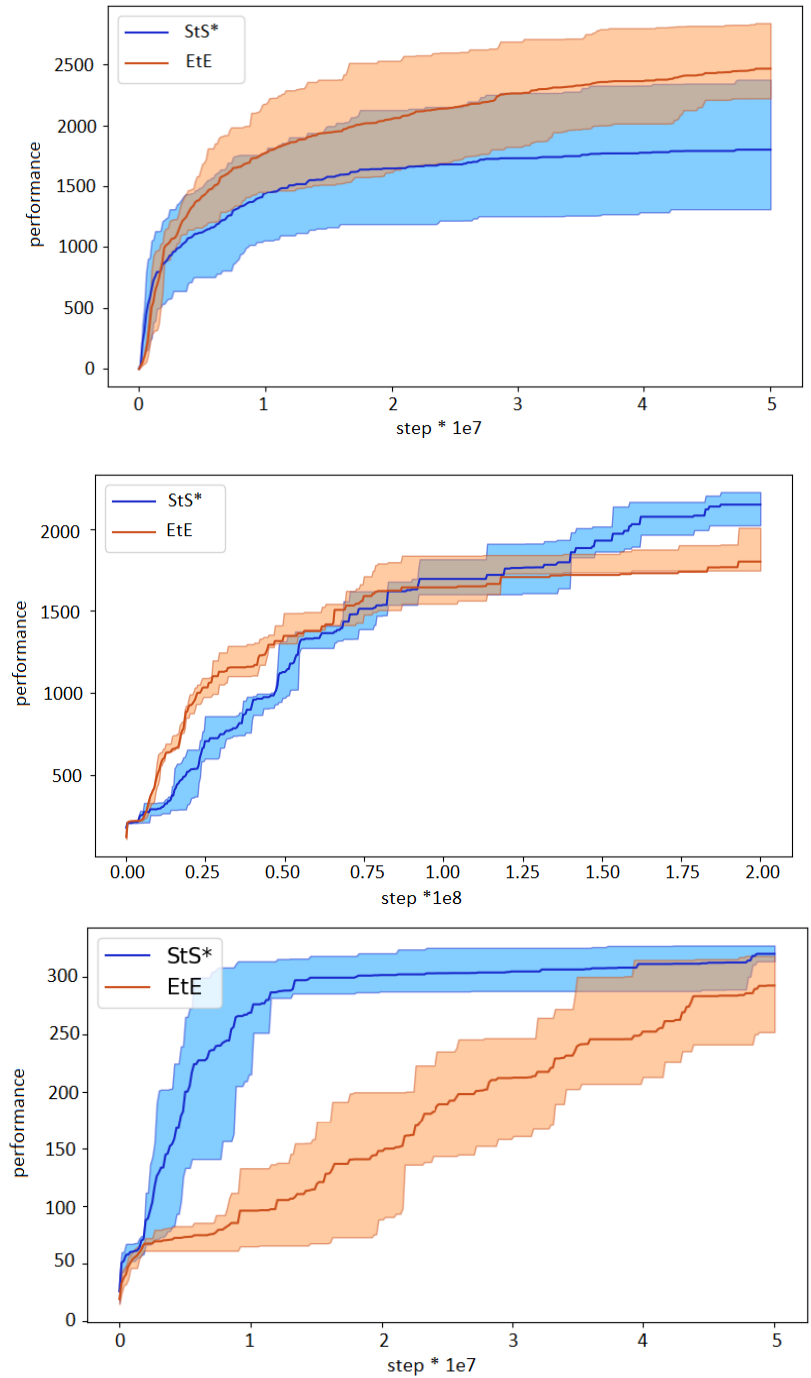

Figure 4. Performance during the training process in the case of the HalfCheetahBullet, BipedalWalkerHardcore, and MIT racecar problems (top, middle and bottom figures, respectively). Performance refer to the average fitness of the best agent achieved so far. Data computed by post-evaluating the best individual of each generation for 3 episodes in the case of MIT racecar and HalfCheetahBullet and for 20 episodes in the case of the BipedalWalkerHardcore. Mean and 90% bootstrapped confidence intervals of the mean (shadow area) across 10 replications per experiment.

## 5. Discussion

The efficacy of evolutionary or reinforcement learning algorithms can be improved by combining the policy network with one or more neural networks trained to extract abstract features from observations through self-supervised methods. Indeed, previous works demonstrated how combined models of this type can speed-up learning and/or achieve better performance also in continuous problems domains. In particular, the research reported in (Riedmiller & VoigtHinder, 2012; Mattner, Lange & Riedmiller, 2012; Ha & Schmidhuber, 2018) demonstrated how the addition of feature-extraction network is beneficial, at least in the case of problems that can benefit from dimensionality reduction and that involve a perspective transformation of the observation states.

In this paper we report new data that provide further evidences on the utility of feature extractions, permit to compare the relative efficacy of alternative methods, and demonstrate the importance of updating the feature extracted during the training of the policy network.

The data reported further support the hypothesis that feature extraction can enhance learning, also in the case of continuous problem domains in which relevant features extend over space and time. Indeed, the usage of feature extraction enabled us to obtain significantly better results in 3 of the 4 problems considered. The utilization of problems that involve agents operating on the basis of egocentric information, instead of allocentric information as in previous studies, demonstrates that feature extraction can be advantageous in general terms, irrespectively from the necessity to perform a perspective transformation. Moreover, the utilization of problems that involve relatively compact observation vectors, instead than large observation vectors as in previous studies, demonstrates that feature extraction can be advantageous also in problems that do not benefit from dimensionality reduction.

The data collected by training the feature extracting network before the policy network, as in previous studies, or also during the training of the policy network demonstrates that the latter technique is much more effective and that the method proposed in this paper for realizing the continuous training is sound.

Finally, the comparison of different self-supervised techniques for extracting useful features demonstrates that sequence-to-sequence learning produces the best results and outperform the other methods used in previous studies in the problem considered.